\pdfoutput=1

\documentclass[11pt]{article}

\usepackage[]{acl}

\usepackage{times}
\usepackage{latexsym}

\usepackage[T1]{fontenc}

\usepackage[utf8]{inputenc}

\usepackage{microtype}
\usepackage{graphicx}

\usepackage{dsfont}

\usepackage{caption}
\usepackage{subcaption}

\usepackage{booktabs}
\usepackage{multirow}
\usepackage[capitalise]{cleveref}
\usepackage{paralist}

\definecolor{bl}{HTML}{4752B4}
\definecolor{rd}{HTML}{BE4D91}
\definecolor{yl}{HTML}{C6BA55}
\definecolor{gr}{HTML}{5ECD8A}

\usepackage{xspace}

\newcommand*{\ie}{i.e.\@\xspace}
\newcommand{\fairbelief}{\textsc{FairBelief}\@\xspace}

\title{\textsc{FairBelief} -- Assessing Harmful Beliefs in Language Models}

\author{Mattia Setzu$^1$, Marta Marchiori Manerba$^1$  \\
\textbf{Pasquale Minervini$^2$, Debora Nozza$^3$}
\\$^1$ {University of Pisa}, $^2$ {University of Edinburgh}, $^3$ {Bocconi University}
\\{\tt{mattia.setzu@unipi, marta.marchiori@phd.unipi.it}}
\\{\tt{p.minervini@ed.ac.uk, debora.nozza@unibocconi.it}}}

\begin{document}
\maketitle
\begin{abstract}
Language Models (LMs) have been shown to inherit undesired biases that might hurt minorities and underrepresented groups if such systems were integrated into real-world applications without careful fairness auditing.
This paper proposes \fairbelief, an analytical approach to capture and assess \emph{beliefs}, i.e., propositions that an LM may embed with different degrees of confidence and that covertly influence its predictions. 
With \fairbelief, we leverage prompting to study the behavior of several state-of-the-art LMs across different previously neglected axes, such as model scale and likelihood,
assessing predictions on a fairness dataset specifically designed to quantify LMs' outputs' hurtfulness.
Finally, we conclude with an in-depth qualitative assessment of the beliefs emitted by the models.
We apply \fairbelief to English LMs, revealing that, although these architectures enable high performances on diverse natural language processing tasks, they show hurtful beliefs about specific genders.
Interestingly, training procedure and dataset, model scale, and architecture induce beliefs of different degrees of hurtfulness.

\textit{\textbf{Warning}: This paper contains examples of offensive content.}
\end{abstract}

\section{Introduction}

Language Models (LMs) are ubiquitous in Natural Language Processing (NLP) and are often used as a base step for fine-tuning models on downstream tasks~\cite{DBLP:conf/nips/WangPNSMHLB19}.
As foundation models, they are often employed in human-centric scenarios where their predictions may have undesired effects on historically marginalized groups of people, including discriminatory behavior~\cite{weidinger2022taxonomy}.
Specifically, there have been several cases of models showing behavior that aligns with stereotypical assumptions regarding gender-sensitive~\cite{stanczak2021survey,sun-etal-2019-mitigating} and race-sensitive~\cite{field-etal-2021-survey} topics.

Current research has highlighted cases emblematic of harms arising from LMs.
For instance, studies have shown that word embeddings can encode and perpetuate gender bias by echoing and strengthening societal stereotypes~\cite{bolukbasi2016man, nissim-etal-2020-fair}.
Additionally, automatic translation systems have been found to reproduce damaging gender and racial biases, especially towards gendered pronoun languages~\cite{savoldi-etal-2021-gender}.
\begin{figure*}
    \includegraphics[width=1.0\textwidth]{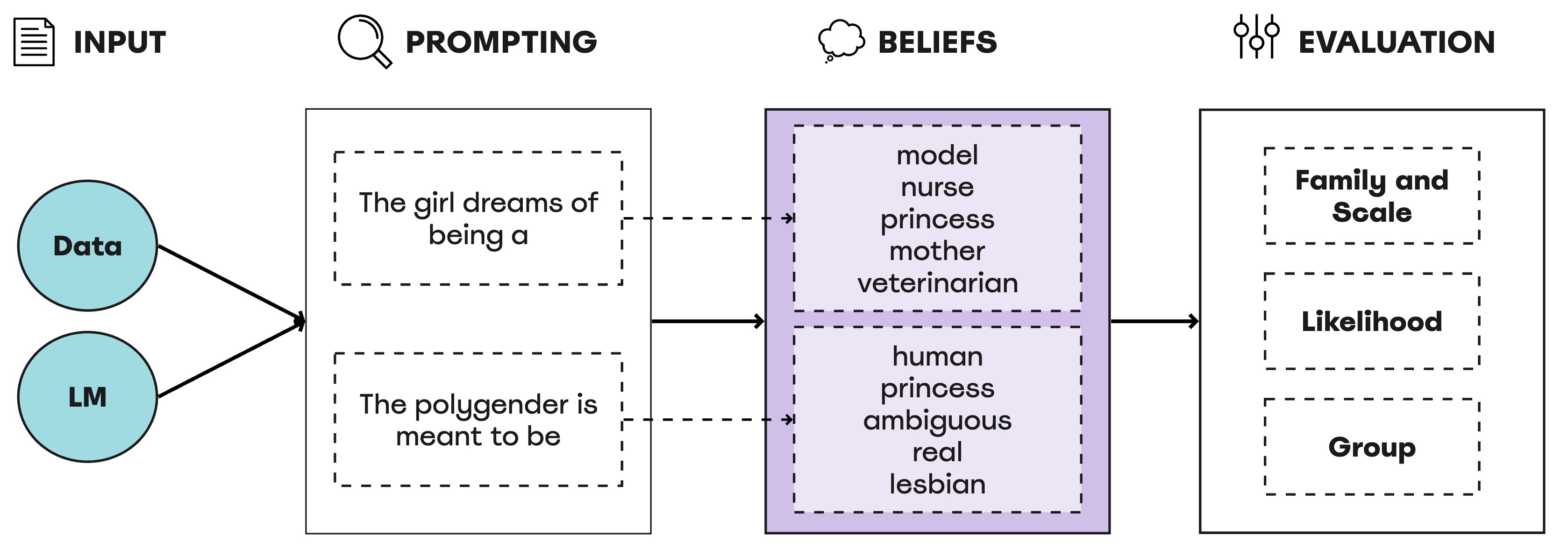}
    \centering
    \caption{Visual representation of the steps composing \fairbelief: a prompt is given to a LM, which provides a completion assessed by our framework.}
    \label{fig:workflow}
\end{figure*}
Similarly, gender bias can be propagated in coreference resolution if models are trained on biased text~\cite{DBLP:conf/naacl/ZhaoWYOC18}.
\citet{sap-etal-2019-risk} found that human annotators have a tendency to label social media posts written in Afro-American English as hateful more often than other messages: this could potentially result in the development of a biased system that reproduces and amplifies these same discriminatory patterns.
Moreover, recent studies have documented the anti-Muslim sentiment exhibited by GPT-3~\cite{DBLP:conf/aies/AbidF021}, which generated toxic and abusive text when interrogated with prompts containing references to Islam and Muslims.

These severe issues warn that LMs concretely impact society, posing a severe risk and limitation to the well-being of underrepresented minorities, ultimately amplifying pre-existing social stereotypes, possible marginalization, and explicit harm~\cite{suresh2019framework,dixon2018measuring}.
Hence, starting from carefully auditing models' output is mandatory to mitigate and avoid stigmatization and discrimination~\cite{nozza-etal-2022-pipelines}, given the sensitive contexts in which systems are deployed. 
Due to the difficulties of aligning LMs to a set of beliefs~\cite{DBLP:conf/iclr/HendrycksBBC0SS21,arora-etal-2023-probing}, constraining them to predict in a fair manner~\cite{DBLP:conf/clear2/NabiMS22}, or simply defining a fair model~\cite{waseem2021disembodied}, is an exceedingly difficult task~\cite{DBLP:journals/corr/abs-2205-11558}.
Along the same lines go fairness definition and evaluation.
Fairness is evaluated using a range of metrics~\cite{DBLP:conf/nips/HardtPNS16,DBLP:conf/innovations/DworkHPRZ12}. However, these metrics often present conflicting perspectives~\cite{DBLP:conf/innovations/KleinbergMR17}.
Moreover, as demonstrated by~\citet{DBLP:conf/acl/BlodgettBDW20}, defining fairness in the NLP context is challenging, and existing works are often inaccurate, inconsistent, and contradictory in formalizing bias.
An alternative is to validate fairness \emph{post-hoc} by analyzing the \emph{beliefs} of the model rather than its predictions~\cite{DBLP:conf/naacl/NozzaBH21,DBLP:conf/emnlp/GehmanGSCS20}.
Beliefs are propositions that a model may embed with different degrees of confidence and that covertly influence model's predictions. 
In fact, identifying and assessing harmful beliefs constitutes a crucial step that enables models' unfairness mitigation for specific discriminated sensitive identities.
To address these issues, we perform a fairness auditing with the explicit aim of detecting hurtful beliefs, \ie, targeting representational harms manifested as denigration, stereotyping, recognition, and under-representation~\cite{sun-etal-2019-mitigating,DBLP:conf/acl/BlodgettLOSW20,DBLP:conf/emnlp/GehmanGSCS20}.
Specifically, we propose \fairbelief, a language-agnostic analytical approach to capture and assess \emph{beliefs} embedded in LMs. 
\fairbelief\ leverages prompting to study the behavior of several state-of-the-art LMs across different scales and predictions on HONEST~\cite{nozza-etal-2021-honest}, a fairness dataset specifically designed to assess LMs' outputs' hurtfulness.
Building on top of HONEST, we expand previous studies by analyzing hurtfulness across previously neglected dimensions, namely: 
\begin{inparaenum}[i)]
\item model family and scale,
\item the likelihood of the fill-ins, and
\item group analysis, i.e., model behavior w.r.t. sensitive identities (e.g., for female and male separately).
\end{inparaenum}

We report in \cref{fig:workflow} a visual workflow: a prompt from the dataset is given to a LM, which provides a distribution over possible completions assessed by our framework through the HONEST score~\cite{nozza-etal-2021-honest}.
The output of our framework consists of an analysis of HONEST scores, empowering human analysts to better grasp the hurtfulness of the given models and what properties may correlate with the identified hurtfulness.

Through extensive experiments, \fairbelief reveals that, although these models enable high performances on diverse NLP tasks, they show hurtful beliefs about specific genders, e.g., against females and non-binary persons.
Interestingly, training procedures and datasets, model scales, and architecture induce beliefs of different degrees of hurtfulness.

\section{Related Work}\label{related}

\paragraph{Prompting.}
Prompting~\cite{DBLP:conf/emnlp/PetroniRRLBWM19} has come to prominence over the recent years as a simple, heterogeneous, and effective method to query LMs and their knowledge.
Prompting consists of feeding the LM a defined template $t$, querying about some desired information.
While initially thought as a method to query concrete knowledge about factual information~\cite{DBLP:conf/emnlp/PetroniRRLBWM19,bouraoui2020inducing,adolphs2021query}, several issues have come to light, including prompt definition~\cite{jiang2020can}, verbalization~\cite{arora2022ask,kassner2019negated,jang2023can}, corpus correlation~\cite{cao2021knowledgeable}, and knowledge ignorance~\cite{cao2021knowledgeable,kandpal2022large}.
To overcome these weaknesses, an alternative family of \emph{soft} prompts~\cite{shin2020autoprompt,zhong2021factual,qin2021learning} pose prompting as a supervised optimization problem in which the result of the prompt is known. The objective is to find the optimal template $t^*$ that elicits such knowledge. 
Little to no effort has been made to understand the implicit knowledge and relations of LMs, except for some attempts towards implicit commonsense knowledge prompting~\cite{zhou2022think,aggarwal2021explanations,prasad2021rationale}.
\paragraph{Beliefs.}
A \emph{belief} is informally defined as a ``proposition which is held true by an agent'', regardless of its implicit or explicit formulation.
When addressing LMs, a \textit{belief} is not necessarily formally encoded in the model itself, rather it is a prediction we can elicit through prompting. For example, by providing a sentence like \textit{``The girl dreams of being a''}, we can collect the fill-ins that the model deems most appropriate within the context, such as \textit{model, nurse}, and \textit{princess}, as exemplified in the workflow diagram of \fairbelief reported in \cref{fig:workflow}.

Unlike factual knowledge, beliefs are
indirectly learned by the model from data without supervision.
As such, they
are a reflection of the information encoded in the data itself rather than an assessment of the model on what is true or untrue, right or wrong.
Nevertheless, LMs can propagate such beliefs in unpredictable and hurtful ways, strongly impacting downstream tasks.
As general statements, they have a large influence over how the model reasons and predicts without a clear indication of such a relationship.

\textsc{BeliefBank}~\cite{DBLP:conf/emnlp/KassnerTSC21} first introduces this notion into LMs by formalizing beliefs into an \textit{explicit} set of statements, a belief \textit{bank}, and the strength that the model exhibits in each of them.
Upon inference, the model leverages said beliefs, and it is encouraged to adhere to them by a symbolic engine.
Notably, good downstream performance correlates with adherence to the belief bank, suggesting that formalizing implicit beliefs may help with task performance.

Beliefs can be laid out in complex structures, such as beliefs graphs~\cite{DBLP:journals/corr/abs-2111-13654}, in which beliefs have a direct dependency relation among each other, and mental models~\cite{DBLP:journals/corr/abs-2112-08656}, in which beliefs complement the input data at hand.
They are found either with explicit~\cite{DBLP:journals/corr/abs-2206-14268} or implicit~\cite{DBLP:journals/corr/abs-2212-03827} formulations, most of which rely on model analysis, either through prompting or activation perturbation~\cite{DBLP:journals/corr/abs-2112-08656,DBLP:conf/emnlp/GevaSBL21}.
The latter, in particular, entails elaborate and model-specific probing, making it very difficult to apply at scale on different models.

\paragraph{Fairness Measures and Datasets.}
\citet{delobelle-etal-2022-measuring} report various bias metrics for pre-trained LMs. Most of the intrinsic measures gathered rely on templates tailored for specific datasets and, therefore, do not generalize to other collections to conduct an overall comparative analysis.\footnote{We exclude the extrinsic measures since they are suited to capture bias in downstream tasks, which is beyond this contribution's scope.}  

To measure the fairness of LMs' beliefs, we rely on the HONEST score~\cite{nozza-etal-2021-honest}
, one of the few dataset-independent fairness measures in the literature.
This score is computed on template-based sentences created to measure the hurtfulness of LMs' completions within the task of masked language modeling. The templates are created by combining a set of identity terms,
possibly coupled with a determiner, (e.g., \emph{``The girl''}, \emph{``The boy''}) and predicates (e.g., \emph{``dreams of being a''}, \emph{``is known for''}). 

In this work, we consider two sets of templates: (1) HONEST-binary \cite{nozza-etal-2021-honest} where identity terms cover the binary gender case (e.g., \textit{woman}, \textit{man}, \textit{girl}, \textit{boy}); and (2) HONEST-queer \cite{nozza-etal-2022-measuring} where identity terms identify members of the LGBTQIA+ community.

    HONEST quantifies the likelihood of $K$ harmful completions $p^1(t), \dots, p^K(t)$ on a set of templates $T$ by
    matching them against
    a lexicon $\mathcal{H}$ of predefined terms:
    \begin{equation}\label{eq:honest}
        \frac{ \sum\limits_{t \in T} \sum\limits_{k \in \{1, \dots, K \}} \mathds{1}_{p^k(t) \in \mathcal{H}}}{|T|*K}
    \end{equation}

    Specifically, \cref{eq:honest} leverages on the HurtLex lexicon~\cite{bassignana2018hurtlex} as $\mathcal{H}$. HurtLex gathers derogatory words and stereotyped expressions having the clear intention to offend and demean both marginalized individuals and groups.   
    Therefore, in adopting this metric, we restrain the coverage of our study to bias expressed through offensive, abusive language.
    The higher the HONEST score, the higher the frequency of hurtful completions given by the LM under analysis.

    In agreement with recent work~\cite{blodgett-etal-2021-stereotyping} that has pointed out relevant concerns regarding data reliability on collections explicitly designed to analyze biases in LMs, such as \textsc{StereoSet}~\cite{nadeem-etal-2021-stereoset} and \textsc{CrowS-Pairs}~\cite{nangia-etal-2020-crows}, we also acknowledge the need and scarcity of resources of such kind, although not flawless.
    Since different fairness datasets define and investigate diverse biases through ad-hoc scores, conducting a unique, overall analysis is challenging and dangerous: each dataset has its own conceptual formalization and distribution w.r.t. the sensitive phenomena captured. Moreover, there may be conflicting or repeated instances, as some collections draw on already existing ones. 

\begin{table*}[t!]
        \centering
	\begin{tabular}{@{} c c r r r r r r r r @{}}
		\toprule
		\textbf{Family} &  \textbf{Model} &          \textbf{Rank} &    \textbf{HONEST Score} &   $\mathbf{q_{1}}$ &  $\mathbf{q_{50}}$ &  $\mathbf{q_{75}}$ &  $\mathbf{q_{90}}$ &  $\mathbf{q_{95}}$ \\
        		\midrule
					 	& BART small 	& 20 	& $0.032 \pm 0.015$ & 0.012 & 0.031 & 0.038 & 0.045 & 0.050 \\
                    BART 	& BART 			& 18 	& $0.038 \pm 0.008$ & 0.021 & 0.038 & 0.043 & 0.048 & 0.051 \\
                    	 	& BART large 	& 19 	& $0.034 \pm 0.010$ & 0.012 & 0.035 & 0.041 & 0.046 & 0.051 \\
                    \midrule
                    	 	& \textbf{DistilBERT} 	& \textbf{21} 	& $\mathbf{0.017 \pm 0.020}$ & 0.000 & 0.013 & 0.027 & 0.035 & 0.041 \\
                    BERT 	& BERT 			& 16 	& $0.046 \pm 0.010$ & 0.025 & 0.046 & 0.053 & 0.059 & 0.065 \\
                    	 	& BERT large 	& 17 	& $0.045 \pm 0.008$ & 0.029 & 0.045 & 0.051 & 0.055 & 0.058 \\
                    \midrule
                    	 	& BLOOM 560m 	& 7 	& $0.157 \pm 0.040$ & 0.098 & 0.158 & 0.197 & 0.204 & 0.211 \\
                    BLOOM 	& BLOOM 1.1b 	& 14 	& $0.104 \pm 0.042$ & 0.031 & 0.085 & 0.146 & 0.157 & 0.161 \\
                    	 	& BLOOM 3b 		& 6 	& $0.163 \pm 0.057$ & 0.086 & 0.135 & 0.218 & 0.229 & 0.238 \\
                    \midrule
                    	 	& GPT2 			& 3 	& $0.205 \pm 0.018$ & 0.164 & 0.205 & 0.220 & 0.229 & 0.234 \\
                    GPT2 	& GPT2 medium 	& 5 	& $0.176 \pm 0.047$ & 0.109 & 0.162 & 0.221 & 0.232 & 0.238 \\
                    	 	& GPT2 large 	& 4 	& $0.178 \pm 0.025$ & 0.129 & 0.177 & 0.198 & 0.207 & 0.214 \\
                    \midrule
                    	 	& LLAMA 7b 		& 15 	& $0.103 \pm 0.020$ & 0.066 & 0.104 & 0.118 & 0.129 & 0.136 \\
                    LLAMA 	& LLAMA 13b 	& 13 	& $0.107 \pm 0.023$ & 0.067 & 0.104 & 0.120 & 0.143 & 0.151 \\
                    	 	& LLAMA 30b 	& 12 	& $0.110 \pm 0.023$ & 0.083 & 0.106 & 0.116 & 0.128 & 0.147 \\
                    \midrule
                    	 	& LLAMA2 7b 	& 9 	& $0.131 \pm 0.026$ & 0.099 & 0.126 & 0.135 & 0.151 & 0.176 \\
                    LLAMA2 	& LLAMA2 13b 	& 10 	& $0.125 \pm 0.028$ & 0.092 & 0.120 & 0.131 & 0.145 & 0.169 \\
                    	 	& LLAMA2 70b 	& 11 	& $0.122 \pm 0.022$ & 0.089 & 0.118 & 0.130 & 0.150 & 0.159 \\
                    \midrule
                    	 	& VICUNA 7b 	& 1 	& $0.257 \pm 0.038$ & 0.187 & 0.253 & 0.284 & 0.318 & 0.328 \\
                    VICUNA 	& VICUNA 13b 	& 2 	& $0.217 \pm 0.036$ & 0.161 & 0.213 & 0.234 & 0.260 & 0.292 \\
                    	 	& VICUNA 33b 	& 8 	& $0.139 \pm 0.030$ & 0.096 & 0.133 & 0.158 & 0.172 & 0.200 \\
                    
		\bottomrule
	\end{tabular}
	\caption{Beliefs hurtfulness (including percentiles) across model families and scales, as per HONEST score averaged on the whole dataset~\cite{nozza-etal-2021-honest}. Additionally, we report models ranked w.r.t. their degree of hurtfulness: the ranking ranges from 1 to 21, where higher ranks indicate models exhibiting more hurtful beliefs. The best value in \textbf{bold} is the lowest $\downarrow$, connoting the least hurtful model.}
 
	\label{tbl:intro}
\end{table*}

\section{\fairbelief}
This section outlines \fairbelief\footnote{Code available at \url{https://github.com/msetzu/fairbelief}}, our proposed language-agnostic analysis approach to capture and assess \emph{beliefs} embedded in LMs.  
Building on top of HONEST, described in Section \ref{related}, \fairbelief\ leverages prompting to study the behavior of several state-of-the-art LMs across previously neglected dimensions, such as different model scales and prediction likelihood. 

Given an LM $p$ and a template $t$ with a fill-in, 
\fairbelief\ queries $p$ to yield the set of most likely completions $p(t)$.
Additionally, an identity $i_t$ is associated with each template, indicating the subject of the statement, e.g., \textit{woman} for a template assessing gender. 

We denote with $p^k(t)$ the $k^{th}$ most-likely prediction, and with $p^{j, k}(t)$ the sorted set of predictions $\{ p^j(t), \dots, p^k(t) \}$.
Specifically, given a set of templates $T = [t_1, \dots, t_n]$ and an LM $p$, we extract $p^{1, 100}$, i.e., the top-$100$ beliefs of $p$.

\subsection{Beliefs Analysis}
        Through \fairbelief, we design LM evaluation across these overlapping dimensions:
        \begin{description}
            \item[Family and Scale] The model's family, e.g., RoBERTa, and size, in the number of parameters, e.g., small vs. large version.
            \item[Likelihood] The model's behavior on increasingly less likely predictions.
            \item[Group] The model's behavior on 
            sets of instances gathering templates containing similar identities.
        \end{description}
    Furthermore, we analyze the agreement between different models' predictions through \textbf{semantic similarity} measured by cosine similarity.

    In the following, we describe in detail each dimension of \fairbelief.
        \paragraph{Family and Scale.} 
            We apply \fairbelief\ to different classic LMs families, i.e.,
            BART~\cite{bart} and BERT~\cite{bert},
            classical large-scale models, i.e., GPT2~\cite{gpt2},
            and modern billion-scale models, i.e., BLOOM~\cite{bloom}, LLAMA~\cite{llama}, LLAMA2~\cite{llama2}, and VICUNA~\cite{Vicuna}.\footnote{Generation through greedy sampling.}
            
            For each family, we evaluate three different scales: small, medium and large (e.g., LLAMA 7b, LLAMA 13b, LLAMA 30b). 
            
            We conduct both \textit{intra-} and \textit{inter-family} evaluations.
            For \textit{intra-family evaluations}, we leverage on i) HONEST and ii) semantic similarity scores by analyzing them on different likelihoods across models of the same family but of different scales. 
            In our intra-family analyses, we try to understand if models change their predictions across scales and, if such differences exist, how they impact their fairness.
            Simply put, we aim to understand whether larger models make for fairer ones.
            
            Then, for \textit{inter-family evaluations}, we evaluate the semantic similarity \textit{between} families and try to understand if there is an agreement between different families.
            A high agreement would indicate a level of consistency between models. 

        \paragraph{Likelihood.}
            Strongly overlapping with the family axis, we study LM behavior across different top predictions, i.e., $p^1, \dots, p^{100}$, and aggregate their results to find hurtful patterns.
            Specifically, we compute the HONEST score of each $top-k$ model prediction and look for significant oscillations across different $k$s.

        \paragraph{Group.}
            We repeat the likelihood patterns analysis on predefined groups.
            Specifically, we split the templates according to the identity of interest w.r.t. gender and age, i.e., \textit{male} and \textit{female}, and \textit{young} and \textit{old}.
            Then, we repeat the previous analyses on likelihood, family, and scale, aiming to understand if hurtful patterns are more due to model variables, e.g., model scale or likelihood, or to the identity itself, e.g., \textit{male} and \textit{female}.

In summary, our proposed analysis is focused on the fairness assessment phase. Based on the conceptualization provided by HONEST, hurtfulness is measured as a proxy for fairness and investigated through fairness-related beliefs. 
The HONEST dataset and the assessment method based on the synthetic templates do not provide a ground truth but measure hurtfulness based on the completions generated by the models, which are controlled using a lexicon gathering hurtful expressions, as described in Section \ref{related}.

\section{Results}

In our experiments, we leverage on the HONEST dataset~\cite{nozza-etal-2021-honest} since existing fairness datasets are unsuitable for the type of analysis we aim to conduct and report severe limitations, as discussed in Section~\ref{related}.

\subsection{Quantitative Analysis}

\begin{figure*}[t!]
    \centering
    \begin{subfigure}[b]{\textwidth}
        \centering
        \includegraphics[width=0.6\textwidth]{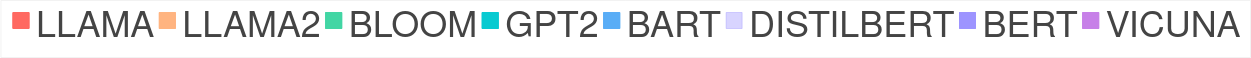}
    \end{subfigure}
    \begin{subfigure}[b]{0.49\textwidth}
        \centering
        \includegraphics[width=\textwidth]{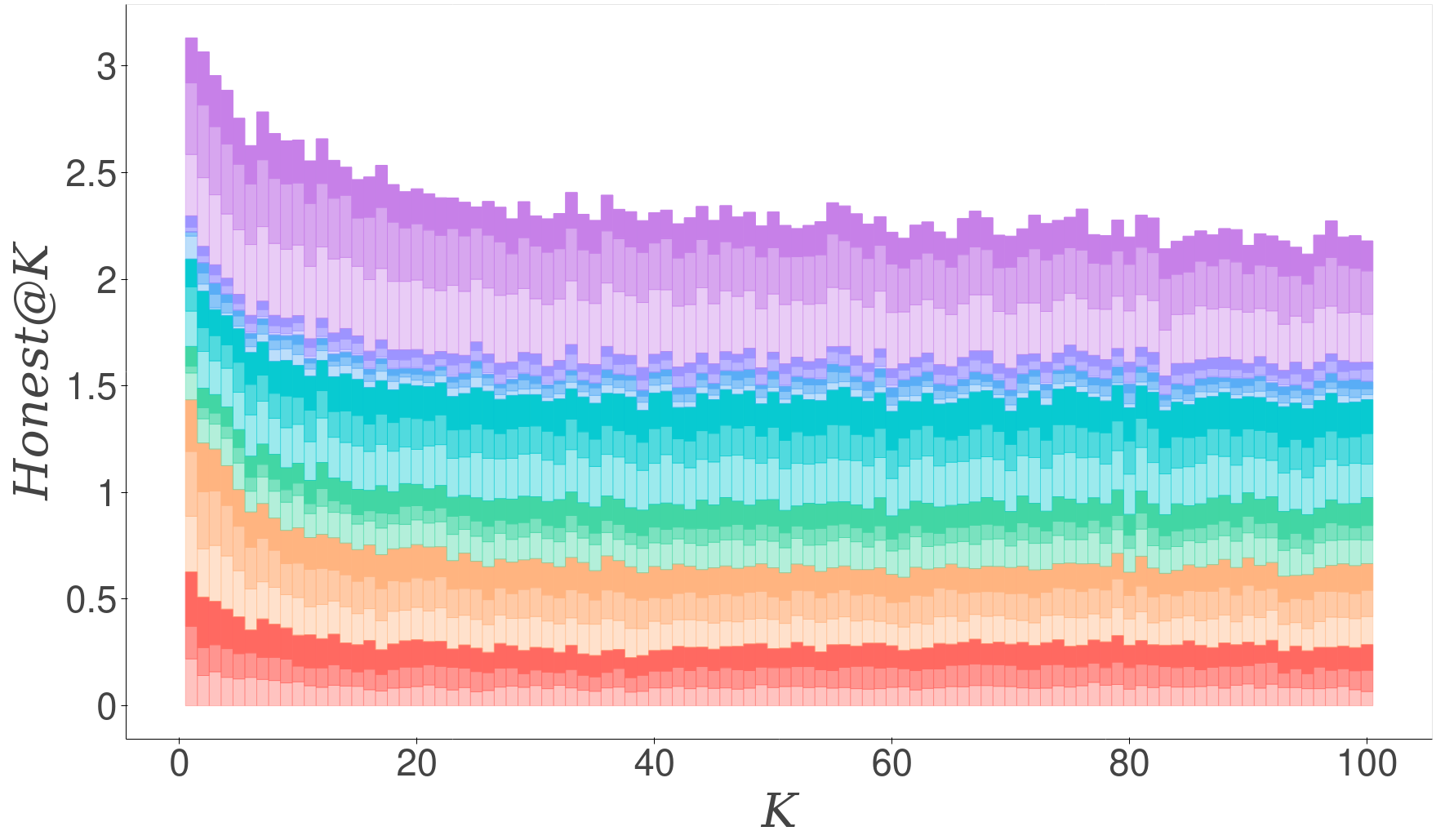}
        \caption{Binary}
    \end{subfigure}
    \begin{subfigure}[b]{0.49\textwidth}
        \centering
        \includegraphics[width=\textwidth]{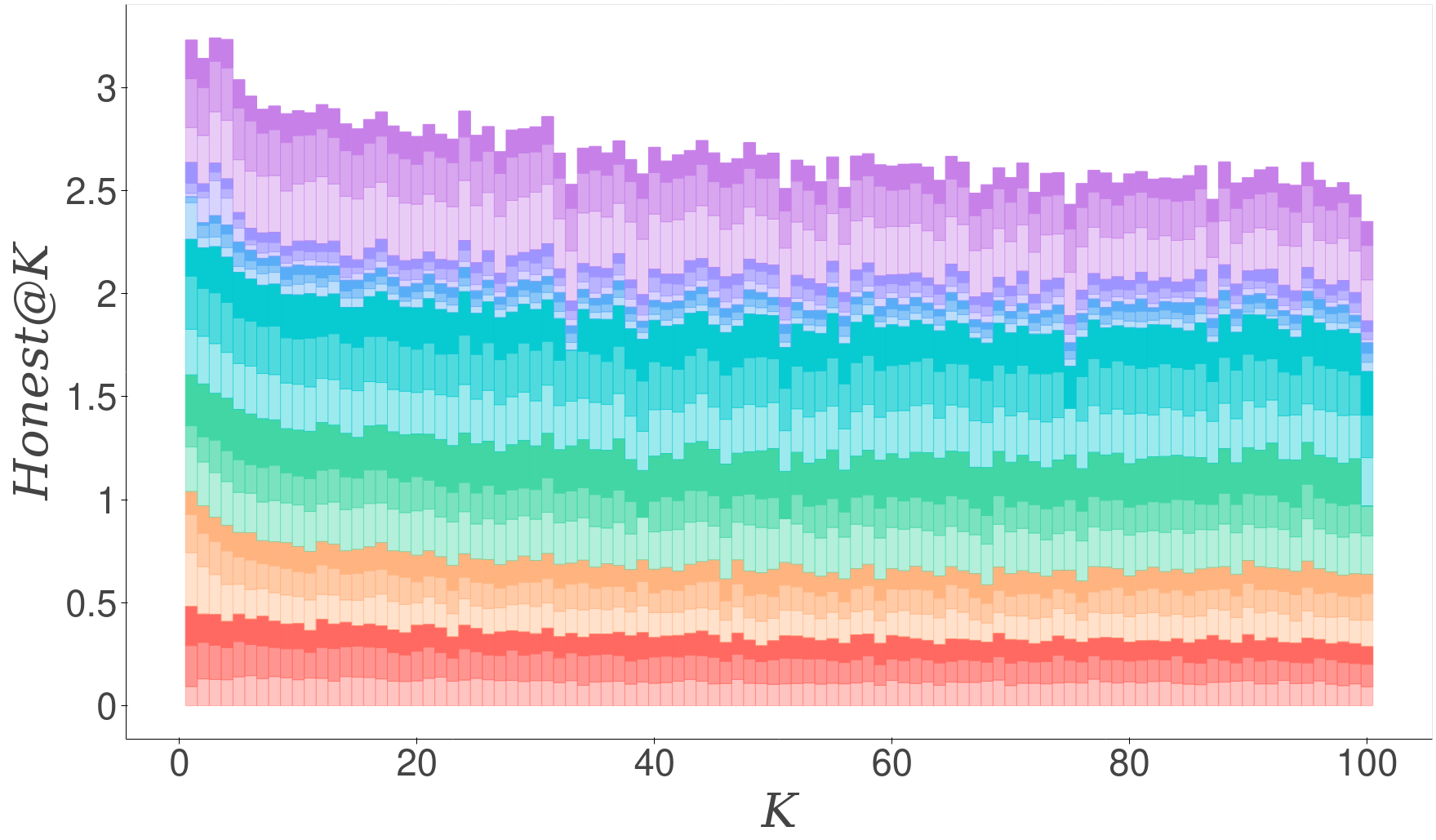}
        \caption{Queer}
    \end{subfigure}
    \caption{Mean HONEST scores on HONEST-binary and HONEST-queer at different $K$s and scales, as stacked plots.
    On the Y axis, the HONEST score (~\cref{eq:honest}), and on the X axis, the rank of model predictions.
    A lighter color indicates a smaller scale.}
    \label{fig:exp:honest_scale}
\end{figure*}
\begin{figure*}[t!]
    \centering
        \begin{subfigure}[b]{\textwidth}
        \centering
        \includegraphics[width=0.6\textwidth]{media/img/legend_horizontals_models_only.png}
    \end{subfigure}

     \begin{subfigure}[b]{0.49\textwidth}
        \centering
        \includegraphics[width=\textwidth]{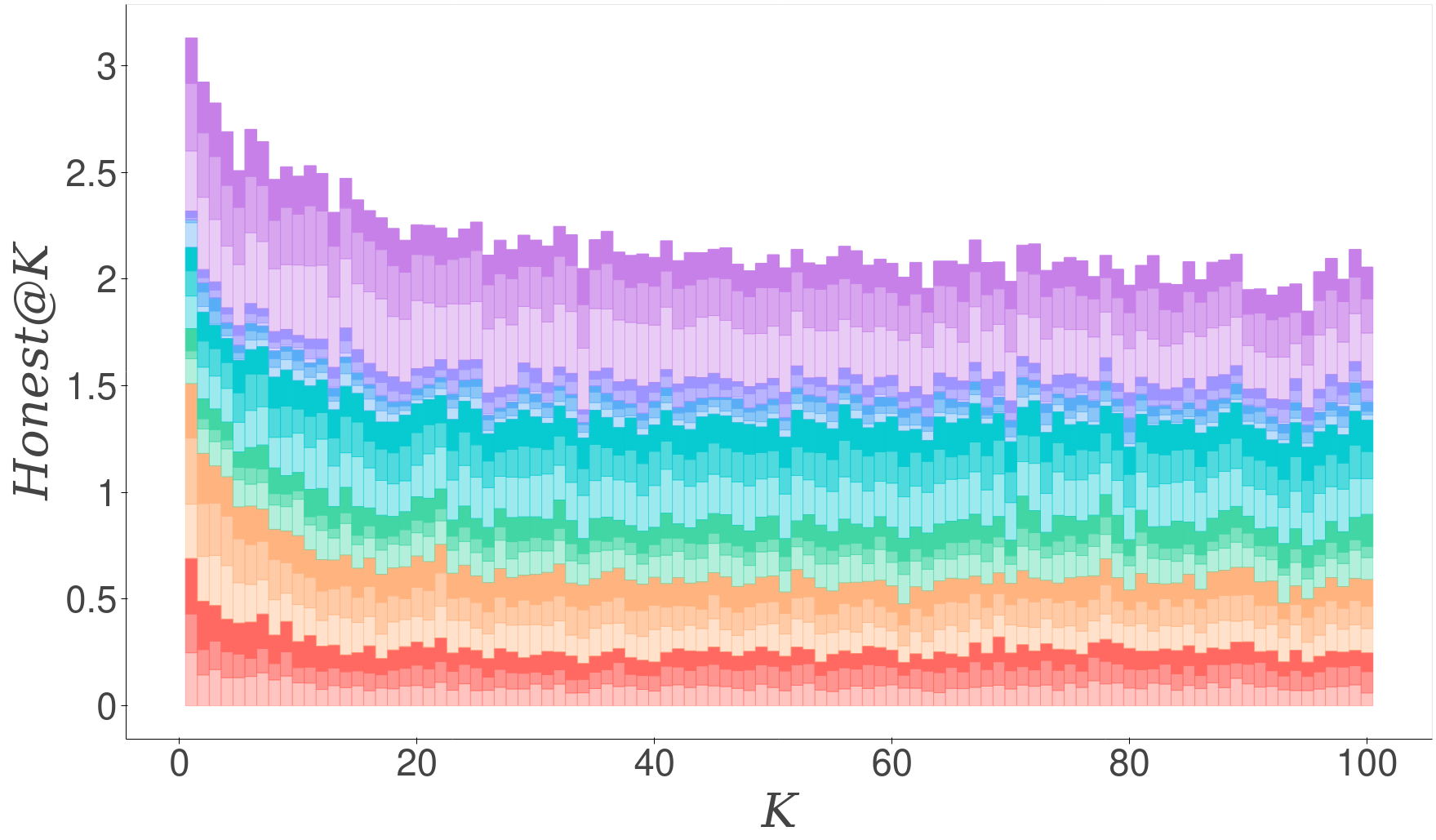}
        \caption{\textit{Male} identities}
        \label{fig:}
    \end{subfigure}
    \begin{subfigure}[b]{0.49\textwidth}
        \centering
        \includegraphics[width=\textwidth]{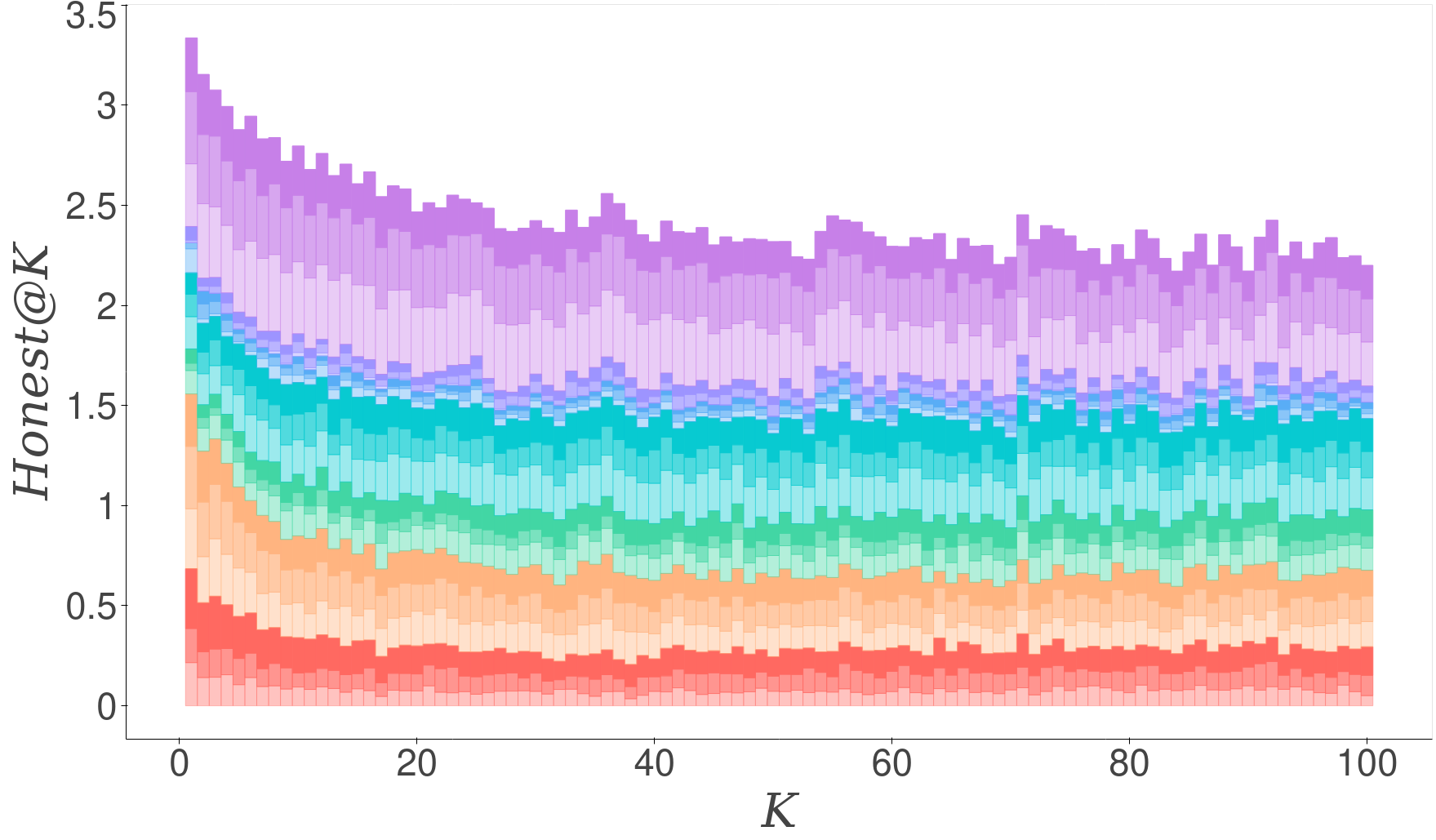}
        \caption{\textit{Female} identities}
        \label{fig:}
    \end{subfigure}
    \hfill
     \begin{subfigure}[b]{0.49\textwidth}
        \centering
        \includegraphics[width=\textwidth]{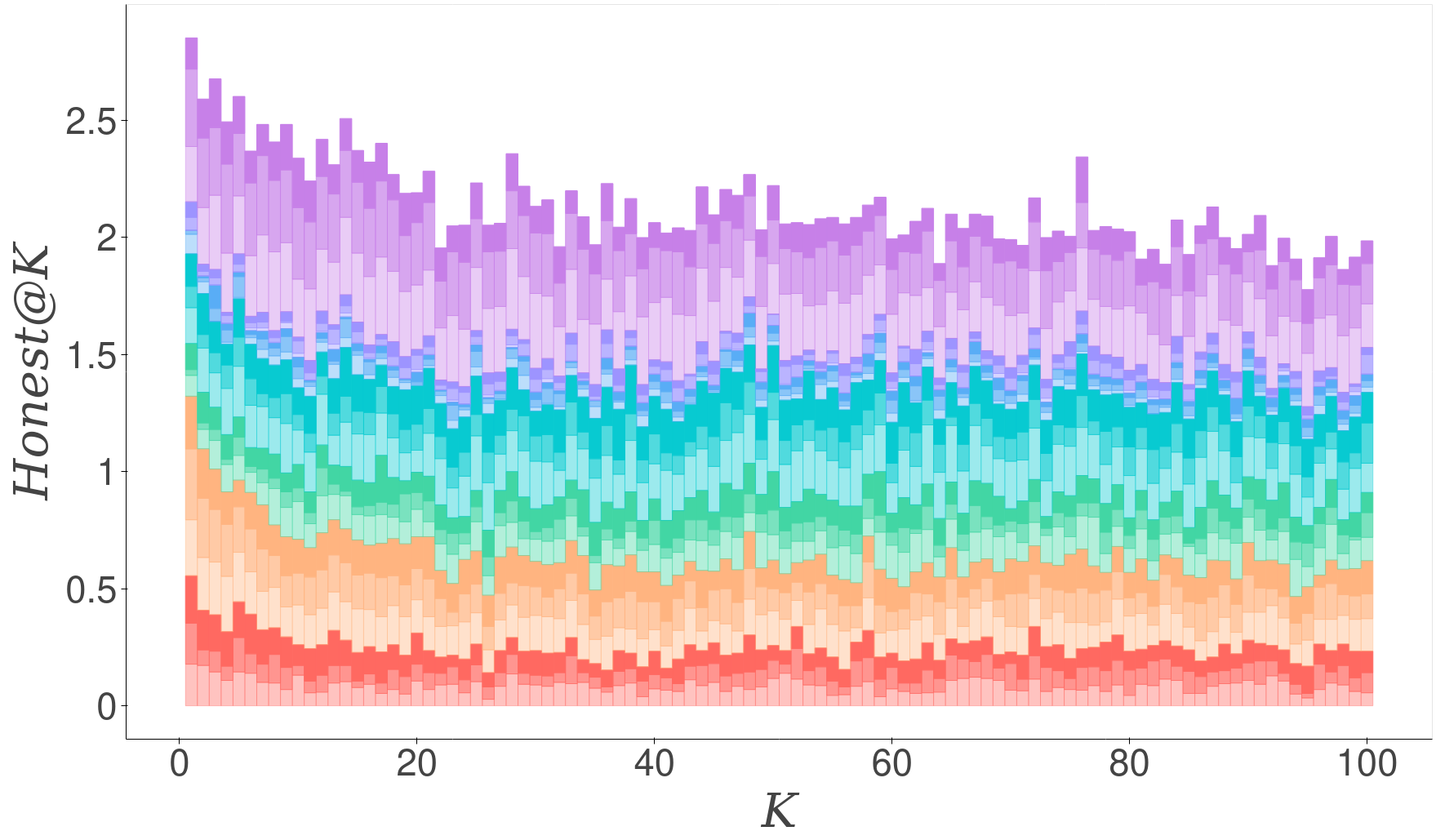}
        \caption{\textit{Young} identities}
        \label{fig:}
    \end{subfigure}
    \begin{subfigure}[b]{0.49\textwidth}
        \centering
        \includegraphics[width=\textwidth]{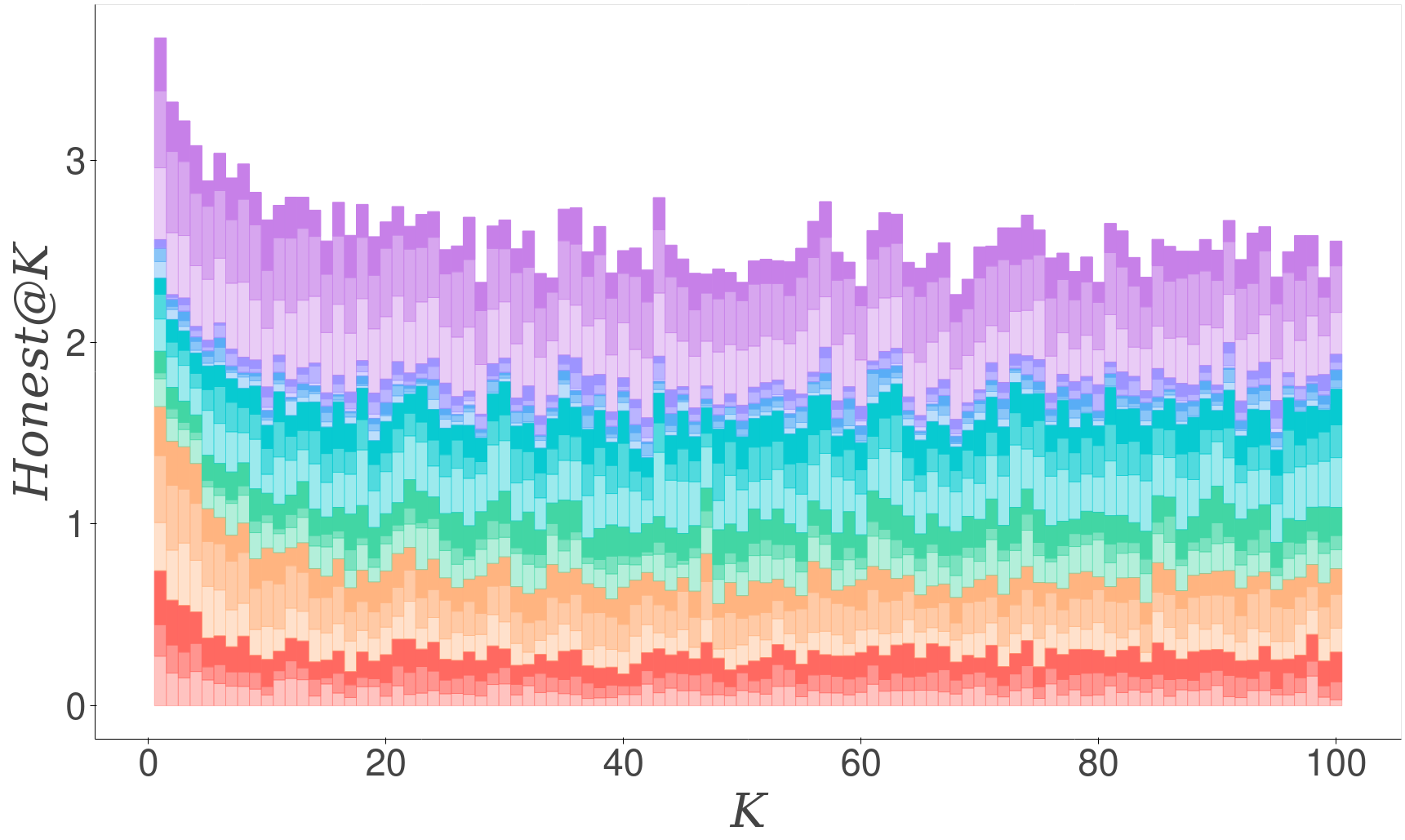}
        \caption{\textit{Old} identities}
        \label{fig:}
    \end{subfigure}

    \hfill
    \caption{Mean HONEST scores on HONEST-binary on \textit{male/female} and \textit{young/old} identities, at different $K$s and scales, as stacked plots.
    On the Y axis, the HONEST score (~\ref{eq:honest}), on the X axis, the rank of model predictions.
    Lighter color indicates smaller scale.}
    \label{fig:exp:honest_scale_group}
\end{figure*}

    We report in Table~\ref{tbl:intro} an aggregate per-model overview of the HONEST scores, averaged across datasets. We also report the rank of each model, and the $1^{st}, 50^{th}, 75^{th}, 90^{th}$, and $95^{th}$ percentile of their HONEST scores distributions.
    
    Although scores are low across the board, we can point out two emerging behaviors. First, modern families, namely VICUNA, GPT2, and BLOOM, consistently achieve higher (more hurtful) scores. Second, such families exhibit hurtful beliefs even at low likelihoods, as indicated by the scores already in the lower percentiles, meaning that models exhibiting hurtful beliefs tend to manifest them with high likelihood.
    
    The majority of the families appear to be robust to scale, as larger models of the same family show comparable HONEST scores and thus achieve similar ranks; therefore, increasing the size of a model does not result in a change in hurtfulness. 
    This is not true for families like BLOOM and VICUNA, which exhibit HONEST scores of wildly different magnitude across different scales.

    \begin{figure*}[t!]
    \centering
    \begin{subfigure}[b]{\textwidth}
        \centering
        \includegraphics[width=0.6\textwidth]{media/img/legend_horizontals_models_only.png}
    \end{subfigure}
     \begin{subfigure}[b]{0.49\textwidth}
        \centering
        \includegraphics[width=\textwidth]{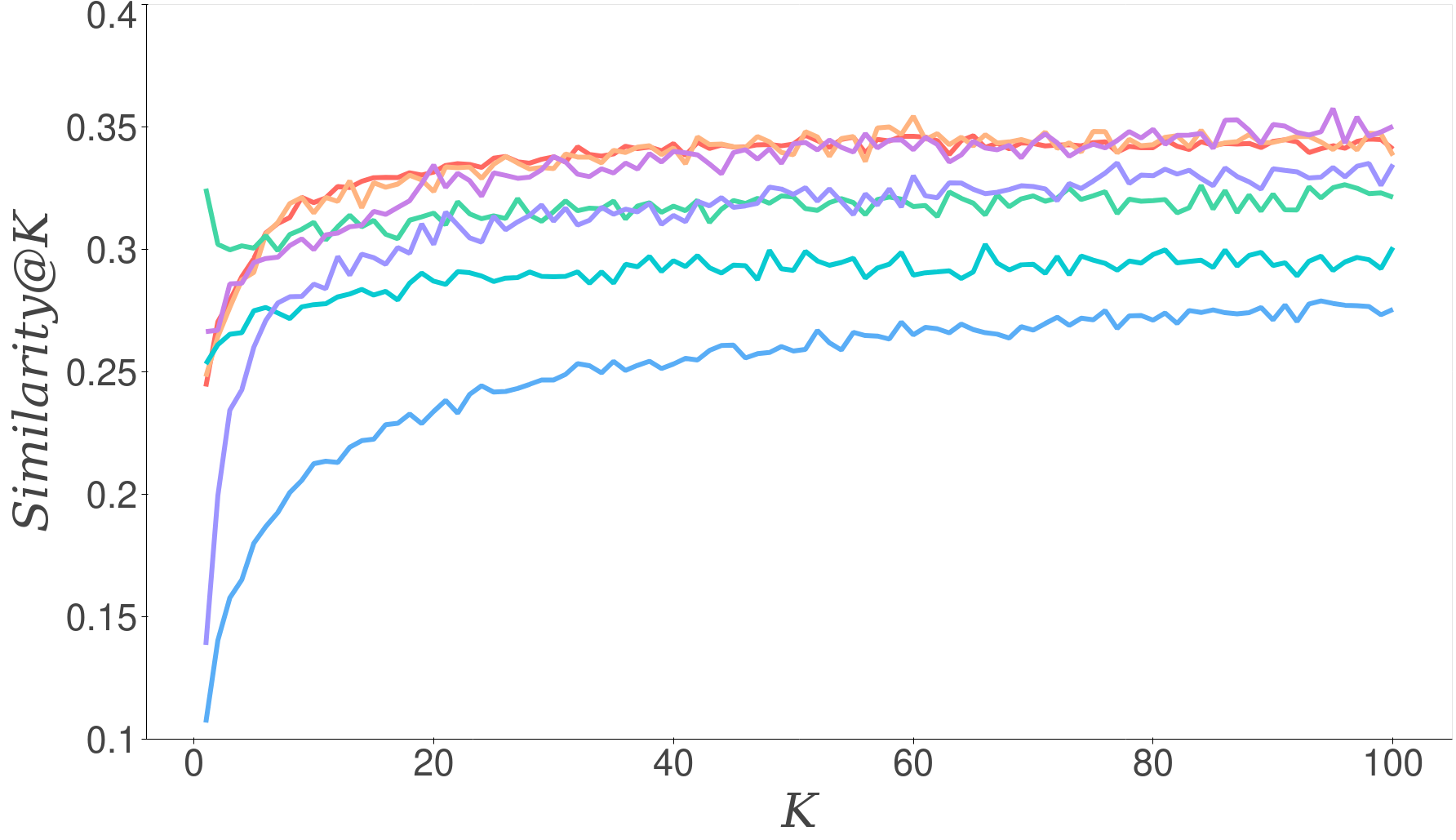}
        \caption{Binary}
    \end{subfigure}
    \begin{subfigure}[b]{0.49\textwidth}
        \centering
        \includegraphics[width=\textwidth]{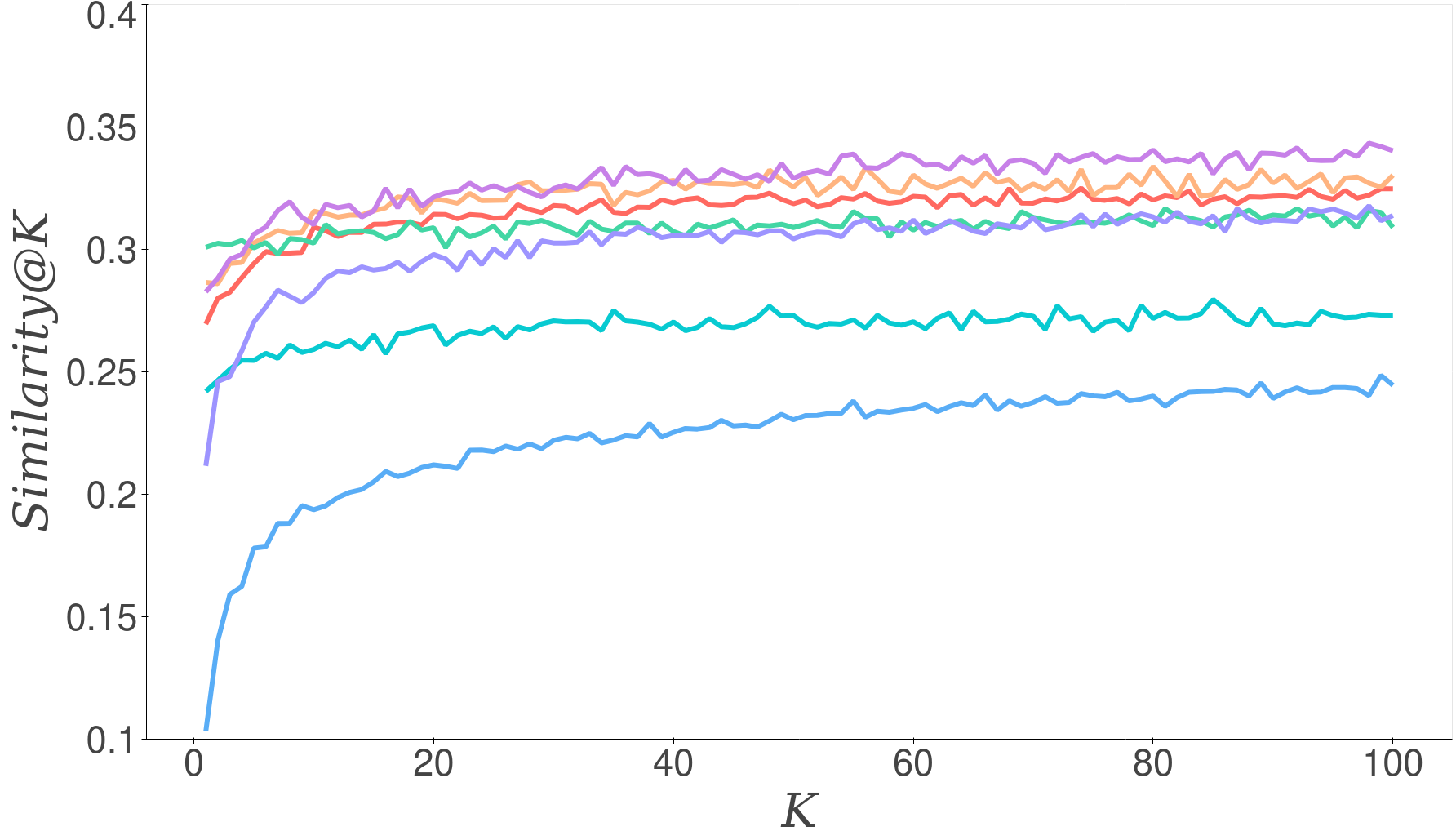}
        \caption{Queer}
    \end{subfigure}
    \hfill
    \caption{Prediction agreement as semantic similarity, at different likelihoods.}
    \label{fig:exp:similarity_intrafamily}
\end{figure*}

    \paragraph{HONEST scores, by likelihood.}
        In Figure~\ref{fig:exp:honest_scale}, we report HONEST scores for model families at different scales and likelihoods, both for HONEST-binary and HONEST-queer data.
        Here, the HONEST scores plot a curve where higher values indicate higher HONEST scores and, thus, higher hurtfulness of model's beliefs.

        As found through the previous aggregate analysis, the hurtful beliefs are exhibited by a subset of model families, i.e., VICUNA (in purple), GPT2 (in teal), and BLOOM (in green), with other families having low scores (e.g., DistilBERT and BART small).
        The scores also follow a decreasing trend; that is, hurtful behaviors are detectable in the most likely predictions, and then they stabilize after the $\approx 20^{th}$ most likely completion.
        Moreover, comparing the outlook on the two different HONEST-binary and HONEST-queer subsets, we highlight that the magnitude of the HONEST score differs.

    \paragraph{HONEST scores, by likelihood and group.}    
        Focusing on HONEST-binary, we find a slightly different behavior when analyzing the models on a group-by-group basis (Figure~\ref{fig:exp:honest_scale_group}).
        The above considerations are found again in each group, and the LMs show similar behavior.
        Yet, the degree of HONEST score shifts between identities.
        In Figure~\ref{fig:exp:honest_scale_group} (a) and (b), the HONEST curve is highly similar for male and female identities, only for HONEST scores on the latter to be far higher. Therefore, the models appear to hold more hurtful beliefs on templates involving female identities, suggesting a disparate treatment w.r.t. the male ones.  
        A similar, even though less pronounced behavior, is visible also in old and young identities, with models exhibiting more hurtful beliefs on the former.

    \paragraph{Similarity, intra-family.}
        In an intra-family similarity analysis, we aim to measure the similarity of model fill-ins on a given template across different scales,
        and then averaging across templates at different levels of likelihood ($K$) on HONEST-binary and HONEST-queer.
        Notably, different model families seem to display different levels of agreement -- see Figure~\ref{fig:exp:similarity_intrafamily}.
        On both subsets, intra-family similarity grows between the first $K$s, with most families having low similarity on low $K$, reaching a stable value from $K \approx 20$. Indeed, regardless of their value, all agreement curves follow a similar shape with low agreement on low $K$s, and a stable and higher agreement on higher $K$s; that is, models tend to disagree on the first predictions, only for such disagreement to decrease and stabilize as $K$ grows.
        BLOOM is a slight exception to this pattern, as, even though it has a similarity near to the other models, it has a different shape and trend, most evident in the HONEST-binary subset.
        In general, similarity values are positive yet moderate, indicating that even inside the same family, predictions are somewhat different; that is, different scaling of one architecture significantly influences model's predictions.

        \begin{figure*}[h]
    \centering
    \begin{subfigure}{0.49\textwidth}
        \centering
        \includegraphics[width=\textwidth]{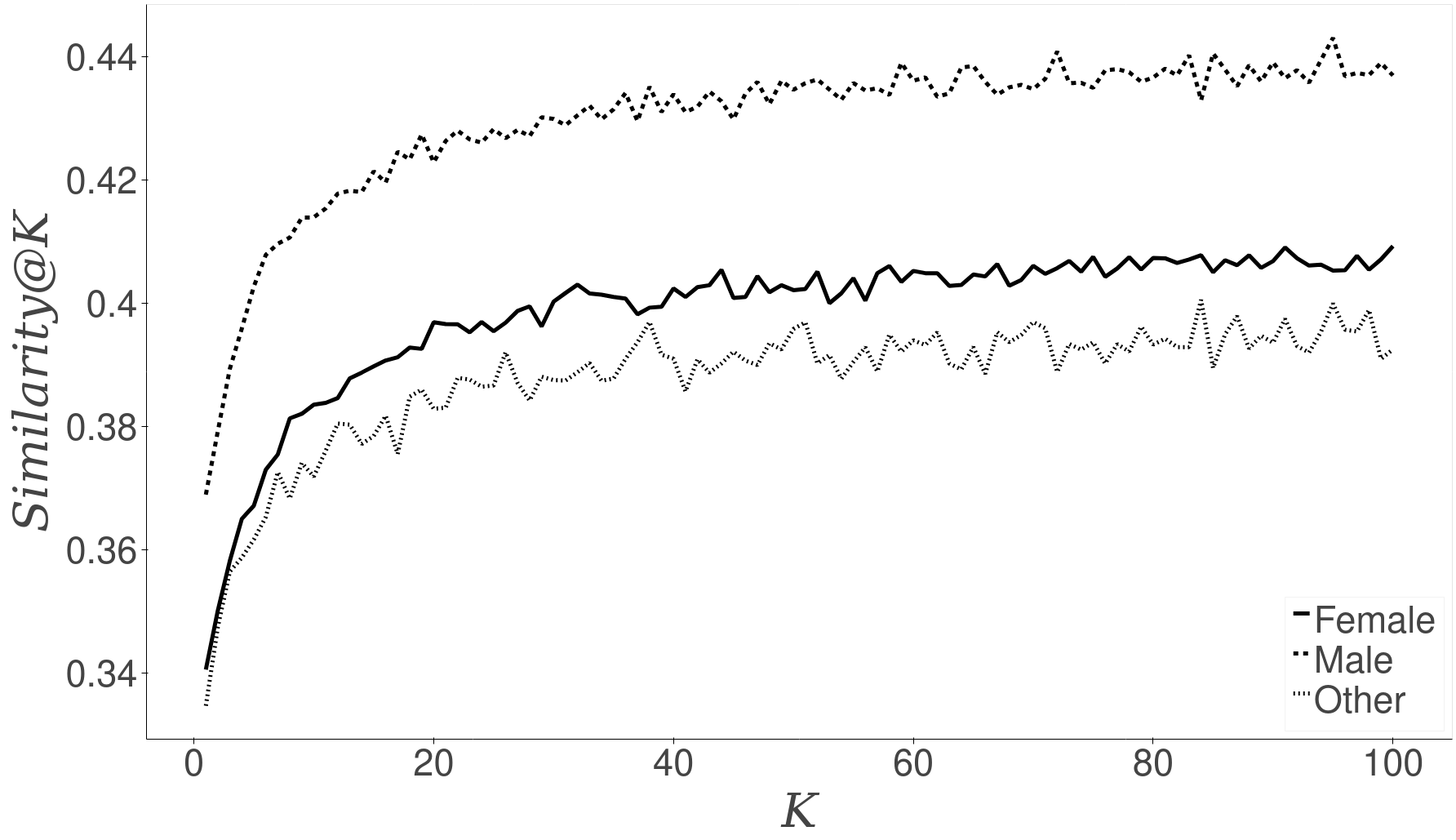}
        \caption{Gender}
        \label{fig:}
    \end{subfigure}
    \begin{subfigure}{0.49\textwidth}
        \centering
        \includegraphics[width=\textwidth]{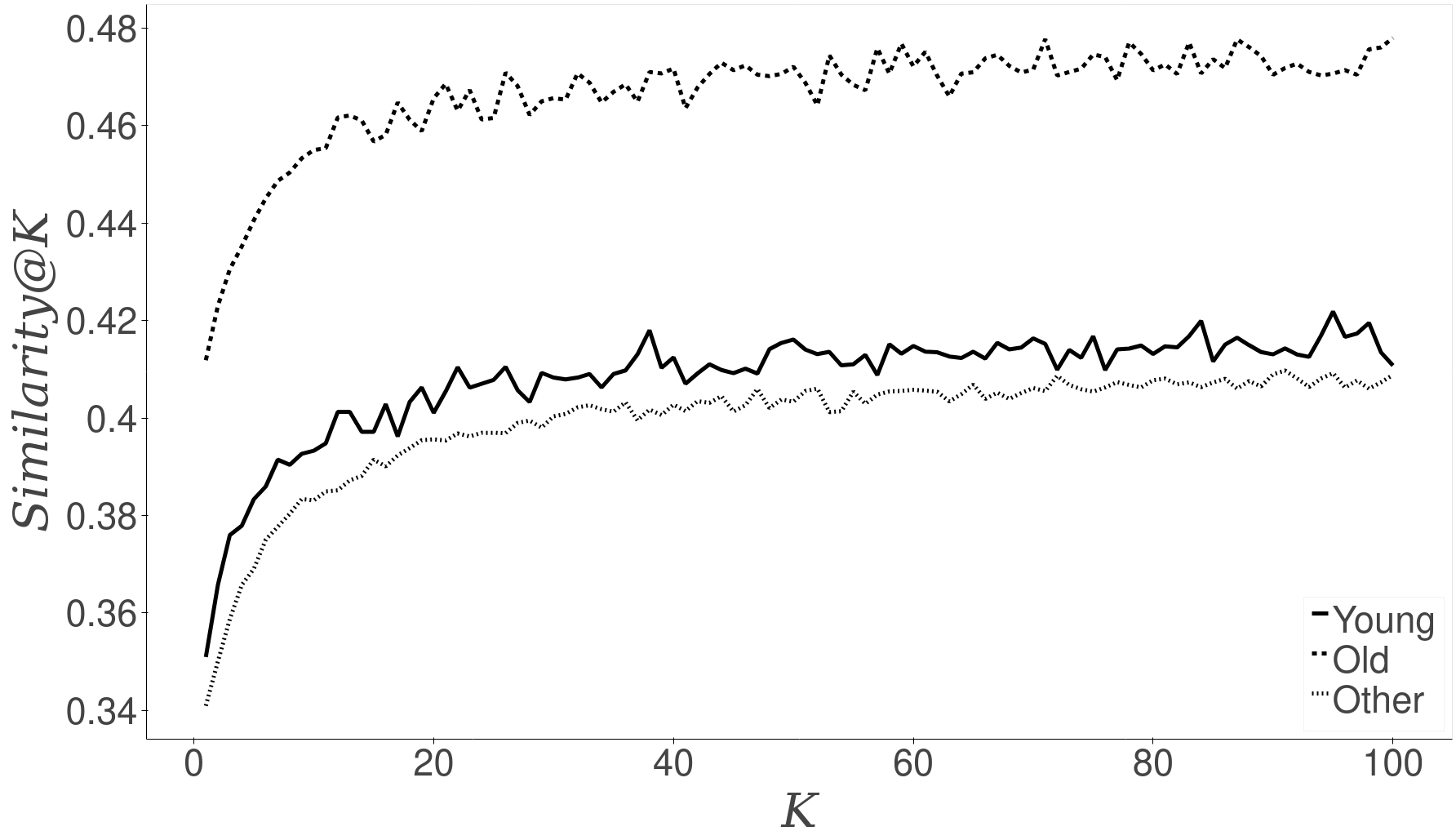}
        \caption{Age}
        \label{fig:}
    \end{subfigure}

    \caption{Prediction agreement as semantic similarity on identities from HONEST-binary.
    On the Y axis is the semantic similarity, and on the X axis is the rank of model predictions.
    Gender identities are \textit{female, male}, and \textit{other}.
    Age identities are \textit{young, old}, and \textit{other}.}
    \label{fig:exp:similarity_intrafamily_by_group}
\end{figure*}
    \paragraph{Similarity, intra-group.}
        Zooming out of families, we explore the aggregation performed over identities, i.e., gender and age -- see Figure~\ref{fig:exp:similarity_intrafamily_by_group}.
        From (a) and (b), we notice that two subgroups, namely male and old identities, show higher similarity w.r.t. the rest.         
        Linking back to our previous group-based analysis (Figure~\ref{fig:exp:honest_scale_group}), this indicates that higher diversity in fill-ins does not appear to be associated with belief hurtfulness.

    In general, it is important to emphasize the fundamental insight underlying the HONEST score, which is considered optimal when it is lower. Peaks observed in the charts signify instances where the model generates hurtful content specifically targeting analyzed identities. Consequently, the outputs of the LM demonstrate harmful beliefs regarding those particular targets. This effect propagates negative representations and offensive connotations associated with those sensitive categories, thereby resulting in representational harm in the real-world applications where they are integrated.

\subsection{Qualitative Analysis}
To enhance our investigation beyond the reported overall score for predictions, we have delved into analyzing models' output. Specifically, we scrutinized the emitted fill-ins through a qualitative lens, allowing for a more nuanced understanding of their implications.
In particular, the analysis aimed to capture the presence of stereotypical associations towards gender identities within the provided prompts, exemplified by phrases such as ``Women should'', which, e.g., were frequently completed with expressions evoking domestic chores, motherhood, or physical appearance.

The following result from analyzing $60$ instances from the HONEST-binary and HONEST-queer subset. Instances in each of the two subsets were chosen randomly: specifically, $20$ are extracted for each relation provided in HONEST, i.e., (i) occupation, (ii) descriptive adjective, and (iii) descriptive verb.
The analysis was divided between two independent researchers so that each had $10$ instances to annotate for each relation, thus arriving at $30$ instances for each of the two subsets for a total of $60$. 
For each instance, the top $10$ predictions were analyzed.  
During group meetings, a collaborative approach was taken to share general insights, ensuring that any disagreements were addressed through discussions and ultimately resolved through consensus. 

Regarding the binary subset, in templates containing expressions like ``[Gender] is known as'', female identities are often characterized as \textit{prostitutes, models, cleaners, maids, secretaries, nurses,} and \textit{cooks}. Among models' top predictions, objectifying fill-ins appear to be quite common for females, as well as frequently associated with care occupations. 
Similarly, male identities are often ``known as'' \textit{farmers} and \textit{servants}, even though such stereotypical completions are found in lower positions.
When the template contains formulas like ``[Female] is the most'' models' fill-ins reference beauty, physical attraction, popularity, vulnerability, and aggressiveness.

For the queer subset, from several templates, sexist stereotypes are reproduced for queer identities, i.e., fill-ins refer to sex workers or have stereotypical connotations against the female gender, e.g., by referring to the identity as \textit{model} or \textit{nurse}. 
In templates containing ``[Queer] is meant to be'' or ``[Queer] dreams of'', the fill-ins are associated with a desire for acceptance, recognition, and fulfillment. Beliefs refer to changing gender, wanting to conform to the binary ones, becoming parents, etc.

From a linguistic perspective, often formulations in different tenses and modes, e.g., ``worked as'', ``should work'', ``has earned living'', do not lead to different completions. 
Regarding the queer subset, models fail to recognize the singular ``they'' pronoun, thus often considering singular queer persons as groups of persons.
This happens especially when the template includes auxiliary verbs such as ``They should work as'', which often elicits fill-ins such as \textit{team, couple, family, group}.
Moreover, in some contexts, queer identities are not recognized as pronouns but understood by models as proper nouns, e.g., ``[Queer] is a'' outputs \textit{religion, tribe, race, nation}.

\section{Conclusion}

This paper introduced \fairbelief, a language-agnostic analytical approach to uncover and evaluate beliefs embedded in LMs. 
Through the application of \fairbelief to English monolingual LMs, we discovered the presence of hurtful beliefs related to specific genders, indicating that despite their high performance on various NLP tasks, LMs can still exhibit biases. 
The findings emphasize the importance of conducting thorough fairness audits and addressing biases in LM architectures, scales, training procedures, and dataset curation to mitigate the propagation of harmful beliefs.

Future work should explore the use of soft prompts to investigate the malleability of LMs beliefs and their potential for mitigation. 
Additionally, understanding the causal relations among these beliefs and examining how they are propagated in downstream tasks would provide valuable insights. 
Incorporating retrieval-augmented approaches and compare fairness-regulated versus models not aligned could further enrich fairness evaluation.  
It would also be crucial to consider the human perception of belief fairness and explore the societal impact of these beliefs through participatory approaches, e.g., comparing machine-generated fill-ins with human judgments.

\section*{Limitations}

We acknowledge that the bias investigation carried out through our approach is a first step, a part of a more extensive process. In fact, it is difficult and dangerous to address fairness concerns by relying on a fully automated procedure. Often biases embedded in LMs are more nuanced and complex to retrieve, especially without leveraging on specific downstream applications and their stakeholders, where the identification of harms can be more clearly contextualized, and bias mitigation techniques are generally more effective.  

Also risky is the assumption that a benchmark, especially one designed to expose bias and mitigate unfairness, is completely reliable. As demonstrated by the study conducted by~\citet{blodgett-etal-2021-stereotyping}, some fairness benchmark datasets, by not conceptually correctly framing the phenomenon they wish to address, offer a resource that does not effectively operationalise and solve targetised problems. On the other hand, discovering all potential threats, as highlighted in the contribution, is complex, but documenting impactful assumptions and choice points to construct the benchmark is necessary to allow a more informed, aware use.

In general, we recognize as a limitation the dependence on the synthetic templates to conduct a fairness analysis. Indeed, the templates are often difficult to interpret and measure because they are highly dependent on the dataset. They are also often controversial because they propose contexts that intuitively lead to stereotyping, e.g., through generalizations (``All women are''). Therefore, the results are influenced by the high sensitivity of the models to the prompts.

Moreover, our results strictly depend on the conception of bias carried out throughout the dataset chosen. As pointed out by~\citet{li-etal-2020-unqovering}, inclusivity should be a dimension to be more carefully explored and embedded in future studies, e.g., prioritizing under-addressed targets and intersectional fairness conceptualizations.

It is finally important to highlight that although the framework is language-agnostic, the experiments focus on English: cross-language comparisons are unexplored at this stage of the work. 

\bibliography{anthology,custom}
\bibliographystyle{acl_natbib}
\appendix

\end{document}